# Plausibility Measures: A User's Guide

**Nir Friedman**
Stanford University
Dept. of Computer Science
Stanford, CA 94305-2140
nir@cs.stanford.edu
http://robotics.stanford.edu/people/nir

**Joseph Y. Halpern**
IBM Almaden Research Center
650 Harry Road
San Jose, CA 95120-6099
halpern@almaden.ibm.com

## Abstract

We examine a new approach to modeling uncertainty based on *plausibility measures*, where a plausibility measure just associates with an event its *plausibility*, an element is some partially ordered set. This approach is easily seen to generalize other approaches to modeling uncertainty, such as probability measures, belief functions, and possibility measures. The lack of structure in a plausibility measure makes it easy for us to add structure on an "as needed" basis, letting us examine what is required to ensure that a plausibility measure has certain properties of interest. This gives us insight into the essential features of the properties in question, while allowing us to prove general results that apply to many approaches to reasoning about uncertainty. Plausibility measures have already proved useful in analyzing default reasoning. In this paper, we examine their "algebraic properties", analogues to the use of $+$ and $\times$ in probability theory. An understanding of such properties will be essential if plausibility measures are to be used in practice as a representation tool.

## 1  INTRODUCTION

We must reason and act in an uncertain world. There may be uncertainty about the state of the world, uncertainty about the effects of our actions, and uncertainty about other agents' actions. The standard approach to modeling uncertainty is probability theory. In recent years, researchers, motivated by varying concerns including a dissatisfaction with some of the axioms of probability and a desire to represent information more qualitatively, have introduced a number of generalizations and alternatives to probability, such as Dempster-Shafer *belief functions* [Shafer 1976], *possibility measures* [Dubois and Prade 1990], and *qualitative probability* [Fine 1973]. Our aim is to examine what is perhaps the most general approach possible to representing uncertainty, which we call a *plausibility measure*. A plausibility measure associates with a set its *plausibility*, which is just an element in a partially ordered space. Every systematic approach for dealing with uncertainty that we are aware of can be viewed as a plausibility measure. Given how little structure we have required of a plausibility measure, this is perhaps not surprising.

As we shall show, this lack of structure turns out to be a significant advantage of plausibility measures. By adding structure on an "as needed" basis, we are able to understand what is required to ensure that a plausibility measure has certain properties of interest. This gives us insight into the essential features of the properties in question while allowing us to prove general results that apply to many approaches to reasoning about uncertainty. For example, in [Friedman and Halpern 1995a] we examine a necessary and sufficient condition for getting the KLM properties [Kraus et al. 1990] for defaults. Once we identify this condition, it is quite easy to show why many different approaches (such as *preferential structures* [Kraus et al. 1990], *$\kappa$-rankings* [Spohn 1987; Goldszmidt and Pearl 1992], and possibility measures [Benferhat et al. 1992]) all satisfy the KLM properties. Moreover, we also describe a weak necessary and sufficient condition for the KLM properties to be complete. This condition is easily seen to hold in all of these approaches. These results help us understand why the KLM properties characterize default reasoning in several different approaches.

In this paper, we move beyond the realm of qualitative and default reasoning, and take a more general look at plausibility measures. If plausibility measures are to be used as a tool for representing uncertainty, then we shall need to be able to emulate some aspects of probabilistic reasoning, such as reasoning by cases and the ability to condition. For example, a standard approach to computing $\Pr(A)$ is to write it as $\Pr(A|B_1)\Pr(B_1) + \cdots + \Pr(A|B_n)\Pr(B_n)$, where $B_1, \ldots, B_n$ is a partition of the space. Other techniques for representing uncertainty, such as $\kappa$-rankings and possibility measures, support similar reasoning. We examine what properties a plausibility measure must satisfy to allow us to carry out such reasoning. We provide constructions that will be useful for any user of plausibility, showing, for example, how we can start with an unconditional plausibility measure and construct a conditional plausibility measure with attractive properties. We also examine what properties of conditional plausibility are required to capture a plausibilistic analogue of Bayesian networks. Recently,

Darwiche and Ginsberg [Darwiche and Ginsberg 1992; Darwiche 1992] and Weydert [1994] have proposed "abstract" formalisms with somewhat similar aims; we show how plausibility measures provide some advantages over both of these approaches.

The rest of this paper is organized as follows. In Section 2 we describe plausibility measures. We investigate their algebraic properties in Section 3, and apply these results in Section 4 to reasoning about independence and qualitative reasoning. We conclude with some discussion in Section 5.

## 2   PLAUSIBILITY MEASURES

A *probability space* is a tuple $(W, \mathcal{F}, \Pr)$, where $W$ is a set of worlds, $\mathcal{F}$ is an algebra of *measurable* subsets of $W$ (that is, a set of subsets closed under finite union and complementation to which we assign probability), and $\Pr$ is a *probability measure*, that is, a function mapping each set in $\mathcal{F}$ to a number in $[0, 1]$ satisfying the well-known Kolmogorov axioms ($\Pr(\emptyset) = 0, \Pr(W) = 1$, and $\Pr(A \cup B) = \Pr(A) + \Pr(B)$ if $A$ and $B$ are disjoint).[1]

A *plausibility space* is a direct generalization of a probability space. We simply replace the probability measure $\Pr$ by a *plausibility measure* Pl, which, rather than mapping sets in $\mathcal{F}$ to numbers in $[0, 1]$, maps them to elements in some arbitrary partially ordered set. We read $\text{Pl}(A)$ as "the plausibility of set $A$". If $\text{Pl}(A) \leq \text{Pl}(B)$, then $B$ is at least as plausible as $A$. Formally, a *plausibility space* is a tuple $S = (W, \mathcal{F}, D, \text{Pl})$, where $W$ is a set of worlds, $\mathcal{F}$ is an algebra of subsets of $W$, $D$ is a domain of *plausibility values* partially ordered by a relation $\leq_D$ (so that $\leq_D$ is reflexive, transitive, and anti-symmetric), and Pl maps the sets in $\mathcal{F}$ to $D$. We assume that $D$ is *pointed*: that is, it contains two special elements $\top_D$ and $\bot_D$ such that $\bot_D \leq_D d \leq_D \top_D$ for all $d \in D$; we further assume that $\text{Pl}(W) = \top_D$ and $\text{Pl}(\emptyset) = \bot_D$. As usual, we define the ordering $<_D$ by taking $d_1 <_D d_2$ if $d_1 \leq_D d_2$ and $d_1 \neq d_2$. We omit the subscript $D$ from $\leq_D, <_D, \top_D,$ and $\bot_D$ whenever it is clear from context.

Some brief remarks on the definition: The algebra $\mathcal{F}$ does not play a significant role in this paper. We have chosen to allow the generality of having an algebra of measurable sets to make it clear that plausibility spaces generalize probability spaces. For ease of exposition, we omit the $\mathcal{F}$ from here on in, always taking it to be $2^W$, and just denote a plausibility space as $(W, D, \text{Pl})$. In some applications we also do not care about the domain $D$. All that matters is the ordering induced by $\leq_D$ on the subsets in $\mathcal{F}$ (see for example [Friedman and Halpern 1995a; Friedman et al. 1995]). However, in dealing with conditional plausibility the domain $D$ plays a more significant role.

So far, the only assumption we have made about plausibility is that $\leq$ is a partial order. We make one further assumption:

**A1.** If $A \subseteq B$, then $\text{Pl}(A) \leq \text{Pl}(B)$.

Thus, a set must be at least as plausible as any of its subsets. While this assumption holds for all the standard approaches to reasoning about uncertainty, we note that there are interesting applications where this might not apply. For example, if we take $\text{Pl}(A)$ to denote how "happy" an agent is if all he knows is that the true world is some world in $A$, then one could well imagine that A1 might not hold. Knowing that the true world is $w$ might be viewed as better than knowing that it is one of $w$ and $w'$. For another example, suppose that we take $\text{Pl}(A)$ to denote the desirability of being in a situation where one needs to choose among elements of $A$. It is well known in the literature on choice theory that agents may occasionally view having more options as a worse situation, not a better one [Kreps 1988]. Despite these caveats, we shall assume A1 for the remainder of the paper.

Clearly plausibility spaces generalize probability spaces. We now briefly discuss a few other notions of uncertainty that they generalize:

- A *belief function Bel* on $W$ is a function $Bel : 2^W \rightarrow [0, 1]$ satisfying certain axioms [Shafer 1976]. These axioms certainly imply property A1, so a belief function is a plausibility measure.

- A *fuzzy measure* (or a *Sugeno measure*) $f$ on $W$ [Wang and Klir 1992] is a function $f : 2^W \mapsto [0, 1]$, that satisfies A1 and some continuity constraints. A *possibility measure* [Dubois and Prade 1990] Poss is a fuzzy measure with the additional property that $\text{Poss}(A) = \sup_{w \in A}(\text{Poss}(\{w\}))$.

- An *ordinal ranking* (or $\kappa$-*ranking*) on $W$ (as defined by [Goldszmidt and Pearl 1992], based on ideas that go back to [Spohn 1987]) is a function $\kappa : 2^W \rightarrow \mathbb{N}^*$, where $\mathbb{N}^* = \mathbb{N} \cup \{\infty\}$, such that $\kappa(W) = 0, \kappa(\emptyset) = \infty$, and $\kappa(A) = \min_{a \in A} \kappa(\{a\})$ if $A \neq \emptyset$. Intuitively, an ordinal ranking assigns a degree of surprise to each subset of worlds in $W$, where 0 means unsurprising and higher numbers denote greater surprise. Again, it is easy to see that if $\kappa$ is a ranking on $W$, then $(W, \mathbb{N}^*, \kappa)$ is a plausibility space, where $x \leq_{\mathbb{N}^*} y$ if and only if $y \leq x$ under the usual ordering on the ordinals.

- A *preference ordering* on $W$ is a partial order $\prec$ over $W$. Intuitively, $w \prec w'$ holds if $w$ is *preferred* to $w'$. Preference orders have been used to provide semantics for *conditional* (or *default*) statements. In [Friedman and Halpern 1995a] we show how to map preference orders on $W$ to plausibility measures on $W$ in a way that preserves the ordering of events of the form $\{w\}$ as well as the truth values of defaults (see Section 4.2 for discussion).

- A *parametrized probability distribution* (PPD) is a tuple $(W, \{\Pr_i : i \geq 0\})$ where each $\Pr_i$ is a probability measure over $W$. Such sequences of measures provide semantics for defaults in $\epsilon$-*semantics* [Pearl 1989; Goldszmidt et al. 1993]. In [Friedman and Halpern 1995a] we show how to map PPDs into plausibility measures in a way that preserves the truth-values of defaults (again, see Section 4.2).

---
[1] Frequently it is also assumed that Pr satisfies *countable additivity*, i.e., if $A_i, i > 0$, are pairwise disjoint, then $\Pr(\bigcup_i A_i) = \sum_i \Pr(A_i)$. We defer a discussion of countable additivity to the full paper.

- A *qualitative probability space* [Savage 1954; Fine 1973] is a pair $(W, \leq^p)$, where $\leq^p$ is a total pre-order over $2^W$ that satisfies several additional properties. Intuitively, $\leq^p$ is a qualitative representation of some probability space Pr such that $A \leq^p B$ if and only if $\Pr(A) \leq \Pr(B)$. A similar notion of *qualitative possibility spaces* is discussed in [Dubois 1986]. It easy to verify that both are instances of plausibility spaces.

For a plausibility measure Pl with a domain $D$ where subtraction makes sense, it is possible to define a dual notion $\text{Pl}^d$ by taking $\text{Pl}^d(A) = \text{Pl}(W) - \text{Pl}(\overline{A})$, where $\overline{A}$ is the complement of $A$. For example, the dual of a belief function is called a *plausibility function* [Shafer 1976] and the dual of a possibility measure is a *necessity measure* [Dubois and Prade 1990]; a probability distribution is its own dual. Note that these dual notions are also plausibility measures.[2] In general, a plausibility measure Pl on $W$ does not have a dual, although it does induce a *dual ordering* $\leq^d$ on subsets of $W$, where $A \leq^d B$ iff $\text{Pl}(\overline{A}) \geq \text{Pl}(\overline{B})$.

Given the simplicity and generality of plausibility measures, we were not surprised to discover that Weber [1991] recently defined a notion of *uncertainty measures*, which is a slight generalization of plausibility measure (in that domains more general than algebras of sets are allowed), and that Greco [1987] defined a notion of $L$-fuzzy measures which is somewhat more restricted than plausibility measures in that the range $D$ is a complete lattice. We expect that others have used variants of this notion as well, although we have not found any further references in the literature. To the best of our knowledge, no systematic investigation of plausibility measures of the type we are initiating here has been carried out before.

## 3 ALGEBRAIC PROPERTIES

In probability theory there is a functional connection—captured by addition—between the probabilities of disjoint sets and the probability of their union. Similarly, there is a functional connection—captured by multiplication—between the conditional probability of $A$ given $B$, the probability of $B$, and the probability of $A \cap B$. These functional connections are frequently used in making probabilistic calculations. Not surprisingly perhaps, many approaches of handling uncertainty have analogues of addition and multiplication with similar roles. In particular, two recent "abstract" approaches to reasoning about uncertainty—that of Darwiche and Ginsberg [Darwiche and Ginsberg 1992; Darwiche 1992] and Weydert [1994]—consider algebras of likelihood values that have some properties of probability, and allow operations analogous to addition and multiplication. As we now show, plausibility measures provide us with the tools to examine the assumptions on plausibility captured by assuming analogues to addition and multiplication. Moreover, the appropriate application of these tools provides us with natural means of going from (unconditional) plausibility to conditional plausibility.

### 3.1 DECOMPOSABLE MEASURES

As we said above, probability theory postulates a functional connection between the probability of disjoint events and the probability of their union. Such an assumption can be viewed as providing a systematic basis for dealing with the combination of evidence, as well as providing a certain modularity in the description of probabilities. For example, if $W$ is finite, a description of the probability of each world in $W$ determines $\Pr(A)$ for any $A \subseteq W$. It is easy to impose a similar requirement on plausibility measures. Consider the following property:

**DECOMP.** If $A$ and $B$ and disjoint, $A'$ and $B'$ are disjoint, $\text{Pl}(A) \leq \text{Pl}(A')$, and $\text{Pl}(B) \leq \text{Pl}(B')$, then $\text{Pl}(A \cup B) \leq \text{Pl}(A' \cup B')$.

We say that a plausibility measure Pl is *decomposable* if it satisfies DECOMP.[3] As we now show, decomposability is enough to force there to be a function $\oplus$ on $D$ such that $\text{Pl}(A \cup B) = \text{Pl}(A) \oplus \text{Pl}(B)$ for disjoint sets $A$ and $B$; we say that $\oplus$ *determines decomposition for* Pl. In fact, the axiom DECOMP= which results from replacing all occurrences of $\leq$ in DECOMP by = is already enough to force there to be a function $\oplus$ that determines decomposition for Pl. DECOMP forces $\oplus$ to have a few additional properties that make it even more like addition.

**Definition 3.1:** Suppose $D$ is a pointed ordered domain and $\circ$ is a partial function mapping $Dom(\circ) \subseteq D \times D$ to $D$. If $t, t'$ are two terms involving $\circ$, we write $t =_e t'$ if $t = t'$ provided $t$ and $t'$ are both defined. (If one of $t$ or $t'$ is not defined, then $t =_e t'$ holds vacuously.) Similarly, we write $t \leq_e t'$ if $t \leq t'$ provided both $t$ and $t'$ are defined. We say that $\circ$ is

- *commutative* if $d_1 \circ d_2 =_e d_2 \circ d_1$,

- *associative* if $(d_1 \circ d_2) \circ d_3 =_e d_1 \circ (d_2 \circ d_3)$

- *monotonic* if $d_1 \leq d_3$ and $d_2 \leq d_4$ implies $d_1 \circ d_2 \leq_e d_3 \circ d_4$,

- *additive* if $d \circ \bot =_e d$, and $d \circ \top =_e \top$,

- *multiplicative* if $d \circ \bot =_e \bot$ and $d \circ \top =_e d$,

- *invertible* if $(d_1, d_2), (d_1, d_3) \in Dom(\oplus)$, $d_1 \circ d_2 \leq d_3 \circ d_4$, and $d_2 \geq d_4 > \bot$ implies $d_1 \leq d_3$.

---

[2]Although the word "plausibility" is used both in our notion of plausibility measure and in the Dempster-Shafer notion of a plausibility function, and plausibility functions are a special case of plausibility measures, there is no other connection between the two notions. There are simply not that many words that can be used to describe notions of uncertainty. We hope the overloading of "plausibility" will not cause confusion.

[3]DECOMP is a weak variant of a property of qualitative probabilities called *disjoint unions* in [Fine 1973, p. 17]. A similar property has been examined in the theory of fuzzy measures [Dubois 1986; Weber 1991].

■

**Theorem 3.2:** *Let $S = (W, D, \text{Pl})$ be a plausibility space. Pl is decomposable if and only if there is a commutative, monotonic, additive function $\oplus$ on $D$ with domain $\text{Dom}(\oplus) = \{(d, d') : \exists A, A' \subseteq W, A \cap A' = \emptyset, \text{Pl}(A) = d, \text{Pl}(A') = d'\}$ that determines decomposition for Pl such that:*

- $\oplus$ *is associative on representations of disjoint sets, i.e., $(d_1 \oplus d_2) \oplus d_3 = d_1 \oplus (d_2 \oplus d_3)$ if there exist pairwise disjoint sets $A_1, A_2, A_3$ such that $\text{Pl}(A_i) = d_i$, $i = 1, 2, 3$.*

Notice that the theorem does not say that $\oplus$ is associative. We show by example in the full paper that, in general, it is not.[4] Interestingly, Fine [1973, p. 22] claimed that associativity of $\oplus$ in his framework follows from the associativity of $\cup$. This claim is not correct, although since there are differences between our assumptions and his, our counterexample does not apply to his framework. It is an open question whether associativity holds in his framework or not.[5] We could, of course, define a (somewhat ugly) condition that would force $\oplus$ to be associative in general. We suspect that this will not be necessary in practice.

Probability measures are decomposable, with decomposition determined by $+$. Similarly, possibility measures and $\kappa$-rankings are decomposable with decomposition determined by max and min, respectively. The embeddings of preferential structures and $\epsilon$-semantics into plausibility structures described in [Friedman and Halpern 1995a] also lead to decomposable plausibility measures. On the other hand, Dempster-Shafer belief functions are not in general decomposable. This follows from the following general observation:

**Lemma 3.3:** *If $(W, D, \text{Pl})$ is a decomposable plausibility space and $A$ and $B$ are disjoint subsets of $W$ such that $\text{Pl}(A) = \text{Pl}(B) = \bot$, then $\text{Pl}(A \cup B) = \bot$.*

**Proof:** Suppose $\text{Pl}(A) = \text{Pl}(B) = \bot$. Since $\text{Pl}(\emptyset) = \bot$, by DECOMP we have $\text{Pl}(A \cup B) = \text{Pl}(A \cup \emptyset) = \text{Pl}(A) = \bot$.
■

Since it is easy to define a belief function *Bel* such that $Bel(A) = Bel(B) = 0$ for two disjoint sets $A$ and $B$, while $Bel(A \cup B) = 1$, it follows that belief functions are not decomposable. A similar argument can be used to show that necessity measures, the duals of possibility measures, are not decomposable in general.

---

[4]That is, we show that there are four pairs of disjoint sets $(A, B), (C, D), (E, F), (G, H)$ such that $\text{Pl}(A) = d_1$, $\text{Pl}(B) = d_2$, $\text{Pl}(C) = d_3$, $\text{Pl}(D) = \text{Pl}(A \cup B) = d_1 \oplus d_2$, $\text{Pl}(E) = d_2$, $\text{Pl}(F) = d_3$, $\text{Pl}(G) = d_1$, $\text{Pl}(H) = \text{Pl}(E \cup F) = d_2 \oplus d_3$, but $\text{Pl}(C \cup D) \ne \text{Pl}(G \cup H)$. Thus, $(d_1 \oplus d_2) \oplus d_3 \ne d_1 \oplus (d_2 \oplus d_3)$, although all terms are defined.

[5]A similar incorrect claim appears in [Darwiche and Ginsberg 1992, p. 623]. However, in [Darwiche 1992], associativity is claimed to hold only for "meaningful sums". This seems to correspond to the same restriction as in Theorem 3.2.

In many cases, we may start with a plausibility defined just on the elements of $W$, not on all subsets of $W$. For example, this may be the case if we try to elicit from the user an ordering on the worlds in $W$, but do not elicit a comparison between sets of worlds. As the next theorem shows, we can then extend this to a decomposable plausibility measure determined by a *total* function $\oplus$ that is commutative, associative, additive, and monotonic.

**Definition 3.4:** We say that the ordered domain $D'$ extends the ordered domain $D$, denoted $D \sqsubseteq D'$, if $D \subseteq D'$, $\bot_D = \bot_{D'}$, $\top_D = \top_{D'}$, and $\le_D$ is $\le_{D'}$ restricted to $D \times D$.
■

**Theorem 3.5:** *Suppose that $D$ is an ordered domain and $\text{pl} : W \to D$. Then there is a decomposable plausibility structure $S = (W, D', \text{Pl})$ such that Pl extends pl (i.e., $D \sqsubseteq D'$ and $\text{Pl}(\{w\}) = \text{pl}(w)$ for $w \in W$) and decomposition for Pl is determined by a total function $\oplus$ that is commutative, associative, additive, and monotonic. Moreover, $S$ is the minimal decomposable extension of pl, in that if $S' = (W, D', \text{Pl}')$ and $\text{Pl}'$ is a decomposable measure that extends pl, then $\text{Pl}(A) \le \text{Pl}(B)$ implies $\text{Pl}'(A) \le \text{Pl}'(B)$ for all subsets $A, B \subseteq W$.*

Thus, if we are given an arbitrary decomposable plausibility space, then the function determining decomposition may not be associative (although it will still be associative in many cases of interest). However, if we are just given a plausibility on elements of $W$, we can *construct* a decomposable plausibility measure determined by a function that is associative. Moreover, the minimality of our construction ensures that it does not make unnecessary assumptions regarding the relative plausibility of sets of worlds.

### 3.2 CONDITIONAL PLAUSIBILITY

Conditioning plays a central role in probabilistic reasoning. Not surprisingly, we are interested in studying conditioning in the context of plausibility as well. Of particular interest will be the connection between the conditional plausibility of $A$ given $B$, and the plausibilities of $B$ and $A \cap B$. Before we can study this relationship, we need to consider conditioning in plausibility structures more generally.

Just as a conditional probability measure associates with each pair of sets $A$ and $B$ a number, usually denoted $\Pr(A|B)$, a conditional plausibility measure associates with pairs of sets a conditional plausibility. Formally, a *conditional plausibility space* is a family $\{(W, D_A, \text{Pl}_A) : A \subseteq W\}$ of plausibility spaces. We typically write $\text{Pl}(B|A)$ rather than $\text{Pl}_A(B)$ and $\text{Pl}(A)$ rather than $\text{Pl}(A|W)$. In keeping with standard practice in probability theory, we also sometimes write $\text{Pl}(B|A, E)$ rather than $\text{Pl}(B|A \cap E)$. Of course, we do not want the various $\text{Pl}_A$'s to be arbitrary. Conditioning attempts to capture the intuition that when we learn $A$, the probability of sets disjoint from $A$ becomes 0, while the relative probability of subsets of $A$ does not change. The following coherence condition guarantees that conditional plausibility spaces have the same property:

**C1.** $\text{Pl}(B|A, E) \le \text{Pl}(C|A, E)$ if and only if $\text{Pl}(A \cap$

$B|E) \leq \text{Pl}(A \cap C|E)$.

In probability theory, the unconditional probability determines the conditional probability, via the relation $\Pr(A|B) = \Pr(A \cap B)/\Pr(B)$. This, of course, is not in general true in arbitrary plausibility measures. We can have two distinct conditional plausibility spaces $\{(W, D_A, \text{Pl}_A) : A \subseteq W\}$ and $\{(W, D'_A, \text{Pl}'_A) : A \subseteq W\}$ on the same space $W$ that agree on the unconditional probability (i.e., $\text{Pl}_W = \text{Pl}'_W$) and yet differ on their components. On the other hand, if all we care about is the ordering of plausibilities, these two conditional plausibility spaces must be essentially the same. To make this precise, we say that two plausibility spaces $(W, D, \text{Pl})$ and $(W, D', \text{Pl}')$ are *(order-)isomorphic* if for any $A, B \subseteq W$, we have that $\text{Pl}(A) \leq_D \text{Pl}(B)$ if and only if $\text{Pl}'(A) \leq_{D'} \text{Pl}'(B)$.

**Proposition 3.6:** *Let $\{(W, D_A, \text{Pl}_A) : A \subseteq W\}$ and $\{(W, D'_A, \text{Pl}'_A) : A \subseteq W\}$ be two conditional plausibility spaces on $W$. If $\text{Pl}_W$ and $\text{Pl}'_W$ are isomorphic, then $\text{Pl}_A$ and $\text{Pl}'_A$ are isomorphic for all $A \subseteq W$.*

Given an (unconditional) plausibility space $S = (W, D, \text{Pl})$, we can find a conditional plausibility space extending $S$ and satisfying C1 in a straightforward way: Consider $\{(W, D_A, \text{Pl}_A) : A \subseteq W\}$, where $D_A$ is a disjoint copy of $\{d \in D : d \leq_D \text{Pl}(A)\}$ and $\text{Pl}_A(B)$ is the element in $D_A$ that corresponds to $\text{Pl}(A \cap B)$. It is easy to see that this conditional plausibility space satisfies C1. Note that since $D_A$ and $D_B$ are disjoint when $A \neq B$, we cannot compare $\text{Pl}(C|A)$ to $\text{Pl}(D|B)$. In fact, it is easy to verify that this plausibility space is the minimal one that extends $S$ and satisfies C1.

In some applications, we want more than just C1. We want there to be a function $\otimes$ such that $\text{Pl}(A \cap B|C) = \text{Pl}(A|B, C) \otimes \text{Pl}(B|C)$, as there is for probability. Such a function $\otimes$ is said to *determine conditioning for* Pl. To study this functional connection (and, more generally, to allow us to compare $\text{Pl}(B|A)$ to $\text{Pl}(B'|A')$ when $A \neq A'$), we consider *standard* conditional plausibility spaces, those for which there is some domain $D$ such that for each $A \subseteq W$, either $D_A = \{\perp_D\}$ or $D_A \subseteq D$.

To force the existence of a function determining conditioning, we require:

**C2.** If $\text{Pl}(A|B, C) \leq \text{Pl}(A'|B', C')$ and $\text{Pl}(B|C) \leq \text{Pl}(B'|C')$, then $\text{Pl}(A \cap B|C) \leq \text{Pl}(A' \cap B'|C')$.

Again, to get the functional dependency, we require only a weaker version of C2 denoted C2$^=$, where all the $\leq$'s are replaced by $=$. Just as with DECOMP, the stronger C2 forces the function $\otimes$ determining conditioning to be monotonic.

Axiom C2 says that $\text{Pl}(A \cap B)$ is determined by $\text{Pl}(A|B)$ and $\text{Pl}(B)$; it does not follow that $\text{Pl}(A|B)$ is determined by $\text{Pl}(A \cap B)$ and $\text{Pl}(B)$. To force this, we need to force $\otimes$ to be "invertible". Roughly speaking, we want a division operator $\ominus$ such that $\text{Pl}(A|B) = \text{Pl}(A \cap B) \ominus \text{Pl}(B)$. Of course, we expect that by increasing the numerator or decreasing the denominator of a fraction like $\text{Pl}(A \cap B) \ominus \text{Pl}(B)$, we get an answer that's at least as large. This motivates the following axiom:

**C3.** If $\text{Pl}(A \cap B|C) \leq \text{Pl}(A' \cap B'|C')$ and $\text{Pl}(B|C) \geq \text{Pl}(B'|C') > \perp$, then $\text{Pl}(A|B, C) \leq \text{Pl}(A'|B', C')$.

Again, we get a weaker version of this axiom, denoted C3$^=$, if we replace all the inequalities with $=$.

**Theorem 3.7:** *Let $S = \{(W, D_A, \text{Pl}_A) : A \subseteq W\}$ be a conditional plausibility space. Pl satisfies C1 and C2 if and only if there exists a multiplicative, monotonic function $\otimes$ with domain $\text{Dom}(\otimes) = \{(d, d') : \exists A, B, C \subseteq W, \text{Pl}(A|B, C) = d, \text{Pl}(B|C) = d'\}$ that determines conditioning for Pl such that:*

- *$\otimes$ satisfies* limited associativity: *if there exist sets $A, B, C, D$ such that $d_1 = \text{Pl}(A|B, C, D)$, $d_2 = \text{Pl}(B|C, D)$, and $d_3 = \text{Pl}(C|D)$, then $(d_1 \otimes d_2) \otimes d_3 = d_1 \otimes (d_2 \otimes d_3)$.*[6]

*Pl additionally satisfies C3 if and only if $\otimes$ is invertible.*

Of course, the standard definition of conditioning in probability satisfies C1–C3, and conditioning is determined by $\times$. Similarly, the standard definition of conditioning in $\kappa$-rankings, which takes $\kappa(A|B) = \kappa(A \cap B) - \kappa(B)$ [Spohn 1987], satisfies C1–C3, with conditioning determined by addition. Finally, the standard notion of conditioning in possibility measures [Dubois and Prade 1990] which takes $\text{Poss}(A|B) = 1$ if $\text{Poss}(A \cap B) = \text{Poss}(B)$ and $\text{Poss}(A|B) = \text{Poss}(A \cap B)$ otherwise, satisfies C1–C3, with conditioning determined by min. It may seem somewhat surprising here that min is invertible, as required by C3. After all, in general, $\min(d_1, d_3) = \min(d_2, d_3)$ does not imply that $d_1 = d_2$. This implication does, however, hold in the domain of min in this case. We remark that we can take an alternate definition of conditioning in possibility theory, by defining $\text{Poss}(A|B) = 0$ if $\text{Poss}(B) = 0$ and $\text{Poss}(A|B) = \text{Poss}(A \cap B)/\text{Poss}(B)$ otherwise. This definition also satisfies C1–C3, with conditioning being determined by multiplication. We shall contrast the two definitions in the next section.

Note that our theorem does not force $\otimes$ to be commutative. It is easy to state conditions that force additional properties of $\otimes$, such as commutativity (see [Fine 1973]).

What happens when we add the requirement of decomposability to conditional plausibility measures? There are actually two ways to do this. We say that a conditional plausibility measure is *locally decomposable* if every $\text{Pl}_A$ is decomposable. This requirement ensures that for each $A$, there is a function $\oplus_A$ such that $\text{Pl}(B \cup C|A) = \text{Pl}(B|A) \oplus_A \text{Pl}(C|A)$ whenever $B$ and $C$ are disjoint. This condition, however,

---

[6]In the full paper we show by example that $\otimes$ is not associative in general. We remark that Fine [1973, p. 30] and Darwiche and Ginsberg [1992, p. 625] again incorrectly claimed that similar assumptions forced $\otimes$ to be associative on all of $\text{Dom}(\otimes)$, and not just on triples $(d_1, d_2, d_3)$ of this special form. Moreover, in this case, the example we provide is a counterexample to associativity also in Fine's framework.

does not relate $\oplus_A$ to $\oplus_B$, i.e., the combination of disjoint events depends on the evidence. We say that conditional plausibility measure is *globally decomposable* (or just decomposable) if it satisfies the following property:

**DECOMP$_c$.** If $A$ and $B$ and disjoint, $A'$ and $B'$ are disjoint, $\text{Pl}(A|C) \leq \text{Pl}(A'|C')$, and $\text{Pl}(B|C) \leq \text{Pl}(B'|C')$, then $\text{Pl}(A \cup B|C) \leq \text{Pl}(A' \cup B'|C')$.

It is easy to state and prove an analogous result to Theorem 3.2, i.e., that a conditional plausibility space is globally decomposable if and only if there is a commutative, monotonic, and additive function $\oplus$ such that for disjoint sets $A$ and $B$, we have $\text{Pl}(A \cup B|C) = \text{Pl}(A|C) \oplus \text{Pl}(B|C)$.

It is not immediately clear that we can find an invertible conditional plausibility extending every (unconditional) plausibility measure. As the next result shows, we can. Moreover, there is a unique minimal algebraic conditional plausibility measure, and it has some very nice properties.

**Theorem 3.8:** *Let $S = (W, D, \text{Pl})$ be a plausibility space. Then there is a conditional plausibility space $S_0 = \{(W, D_0, \text{Pl}_A) : A \subseteq W\}$ extending $S$ (i.e., $D \subseteq D_0$ and $\text{Pl}_W = \text{Pl}$) in which conditioning is determined by a total function $\otimes$ which is commutative, associative, multiplicative, invertible, and monotonic.[7] Moreover, if $S$ is decomposable, then so is $S_0$; in fact, there is a total function $\oplus$ that is commutative, additive, and monotonic that determines decomposition, and $\otimes$ distributes over $\oplus$. In addition, if there is an associative function that determines decomposition for Pl, then $\oplus$ is associative as well.*

The construction of Theorem 3.8 tells us that for any plausibility measure Pl we can find a "nice" (i.e., multiplicative and invertible) conditional plausibility that extends it. Moreover, there is a conditional plausibility measure extending Pl in which the functions determining conditioning and decomposition are total, and have desirable properties including commutativity and associativity of $\otimes$, and distributivity of $\otimes$ over $\oplus$. As we show in the full paper, these properties are not necessarily implied by C1–C3 and DECOMP$_c$ alone. When combined with our construction in Theorem 3.5, this gives us a way of starting with a plausibility measure on worlds in $W$, and extending to a conditional plausibility measure with these attractive properties.

The conditional plausibility measure constructed in Theorem 3.8 is in a precise sense the minimal one with all the required properties. If we start with an unconditional probability measure, $\kappa$ ranking, or possibility measure, the standard approaches to conditioning do not in general give us the conditional measure defined by this construction. Thus, they makes some comparisons between conditional plausibilities that are not forced by our requirements. Interestingly, if we consider the standard approach to conditioning in possibility measures defined earlier (with $\otimes$ being min), it cannot be extended to a total commutative function.

---
[7]Note that the fact that $\otimes$ satisfies these conditions guarantees that $S_0$ satisfies C2 and C3.

**Example 3.9:** Suppose we have a possibility measure Poss on a space $W$ and sets $A \subseteq B \subseteq C$ such that $\text{Poss}(A) = 1/4$, $\text{Poss}(B) = 1/2$, and $\text{Poss}(C) = 3/4$. (It is easy to find such a possibility measure.) Then, according to the definitions, we have $\text{Poss}(A|B,C) = \text{Poss}(A|C) = 1/4$, $\text{Poss}(B|C) = 1/2$, and $\text{Poss}(B|A,C) = 1$. Hence, $\min(\text{Poss}(B|A,C), \text{Poss}(A|C)) = \min(\text{Poss}(A|B,C), \text{Poss}(B|C)) = 1/4$, but $\text{Poss}(A|B) \neq \text{Poss}(C)$. More abstractly, we have $d_1 \otimes d_2 = d_2 \otimes d_3$, but $d_1 \neq d_3$. On the other hand, if $\otimes$ could be extended to a total, commutative, invertible function, we would have to have $d_1 = d_3$. We return to this issue in the next section. ∎

### 3.3 COMPARISON TO OTHER ALGEBRAIC APPROACHES

We now briefly compare our approach to others in the literature.

Darwiche and Ginsberg's [1992] approach is somewhat similar to ours. They start with functions (which they call *states of belief*) that map formulas to an (unordered) set of plausibility values. To relate these to plausibility measures, we can take $W$ to be the set of all truth assignments, and identify a formula with the set of worlds that satisfy it. Darwiche and Ginsberg then describe various assumptions that force the existence of what, in our terminology, are functions $\oplus$ and $\otimes$ that essentially determine decomposition and conditioning. They then define an ordering $\leq_\oplus$ on plausibilities in terms of $\oplus$: $x \leq_\oplus y$ if there is a $z$ such that $x \oplus z = y$. Their assumptions do not force their analogues of $\oplus$ and $\otimes$ to be total (although they are total in all the examples provided).

Notice that Darwiche and Ginsberg assume that originally we have a mapping from worlds to an unordered set of plausibility values. They then impose an ordering on this set. Their construction does not deal so well with the type of situation discussed before Theorem 3.5, where we start with a function pl that assigns to each world a plausibility in some ordered set; there is no way to take into account this initial ordering in their construction. For example, suppose $W = \{a, b\}$ and $\text{pl}(a) > \text{pl}(b)$. Their construction would make $a$ and $b$ incomparable according to $\leq_\oplus$, since there is no $c$ such that $b \oplus c = a$. In particular, this means that the Darwiche-Ginsberg approach cannot deal with initial preferential orders, which arise in default reasoning (see Section 4.2).

Weydert [1994] starts with total functions $\oplus$ and $\otimes$ on some set $W$, where $\oplus$ is commutative, associative, additive, and monotonic, while $\otimes$ is commutative, associative, multiplicative, invertible, and monotonic. (Indeed, he requires even more of $\oplus$ and $\otimes$, although it is beyond the scope of the paper to explain these requirements.) Moreover, he also assumes a totally ordered domain $D$ of plausibilities. He then define plausibility measures (*quasi-measures* in his terminology) as functions from $W$ to $D$ such that $\oplus$ determines decomposition and $\otimes$ determines conditioning.

Our discussion above shows that quasi-measures cannot capture conditional possibility measures, since these cannot be embedded in a total algebra. Similarly, they cannot capture belief functions, since these are not decomposable, nor preferential structures, since these are not totally ordered. Thus, Weydert's framework is not general enough to deal with many of the examples of interest to us.

## 4 APPLICATIONS

In this section, we discuss two possible applications of plausibility: reasoning about independence and qualitative reasoning.

### 4.1 INDEPENDENCE

The notion of independence plays a critical role in reasoning about uncertainty. Intuitively an event $A$ is independent of $B$ given $C$, if, once we know $C$, evidence regarding $B$ does not provide us information about $A$. In probability theory, we say that $A$ and $B$ are independent given $C$ if $\Pr(A|C) = \Pr(A|B \wedge C)$. It is easy to see that this is equivalent to $\Pr(A \wedge B|C) = \Pr(A|C) \times \Pr(B|C)$. Once we consider formalisms other than probability theory, there are a number of alternative definitions of independence that have been considered in the literature; see, for example, [Fine 1973; Goldszmidt and Pearl 1992; Dubios et al. 1994]. We do not consider all the alternatives in our discussion of independence in the context of plausibility structures. Rather, we focus on one possible definition—perhaps the most obvious generalization of the probabilistic definition—and a more qualitative variant of it.

Suppose $A, B, C \subseteq W$. We say that Pl makes $A$ *(plausibilistically) independent of $B$ given $C$* denoted Pl $\models Ind_s(A, B|C)$, if $\text{Pl}(B \cap C) = \bot$ or $\text{Pl}(A|C) = \text{Pl}(A|B, C)$. Thus, $A$ is independent from $B$ given $C$ either if $B$ is implausible or if conditioning on $B$ does not change the plausibility of $A$.

To what extent do properties of probabilistic independence carry over to this definition? That depends on what assumptions we make about the underlying plausibility measure. Once we consider multiplicative plausibility measures (so that it makes sense to use $\otimes$), we can characterize plausibilistic independence much as we did probabilistic independence.

**Proposition 4.1:** *(a) If* Pl *satisfies C2$^=$ and* Pl $\models Ind_s(A, B|C)$, *then* $\text{Pl}(A \cap B|C) = \text{Pl}(A|C) \otimes \text{Pl}(B|C)$.

*(b) If* Pl *satisfies C2$^=$ and C3$^=$, then* Pl $\models Ind_s(A, B|C)$ *if and only if* $(\text{Pl}(A|C), \text{Pl}(B|C)) \in Dom(\otimes)$ *and* $\text{Pl}(A \cap B|C) = \text{Pl}(A|C) \otimes \text{Pl}(B|C)$.

*(c) If* Pl *satisfies DECOMP$_c^=$, $A_1 \cap A_2 = \emptyset$,* Pl $\models Ind_s(A_1, B|C)$, *and* Pl $\models Ind_s(A_2, B|C)$, *then* Pl $\models Ind_s(A_1 \cup A_2, B|C)$.

Thus, if Pl satisfies C2$^=$, then when $A$ and $B$ are unconditionally independent (i.e., independent given $W$), the plausibility of $A \cap B$ is determined by $\text{Pl}(A)$ and $\text{Pl}(B)$. This seems to be a fundamental property of independence, and allows us to modularize the description of Pl. Decomposability gives us another important property of independence: it guarantees that $Ind_s$ is closed under disjoint union.

In probability theory, independence is symmetric: if $A$ is independent of $B$, then $B$ is also independent of $A$. When is $Ind_s$ symmetric? Expanding the definition of $Ind_s$, we see that it is symmetric if whenever $\text{Pl}(A|B, C) = \text{Pl}(A|C)$, then $\text{Pl}(B|A, C) = \text{Pl}(B|C)$. Now if Pl satisfies C2$^=$, then $\text{Pl}(A \cap B|C)$ can be expanded in two ways: $\text{Pl}(A|B, C) \otimes \text{Pl}(B|C)$ and $\text{Pl}(B|A, C) \otimes \text{Pl}(A|C)$. Thus, we would expect $Ind_s$ to be symmetric whenever $\otimes$ is invertible. This may not follow if $\otimes$ is not total and commutative. To see this, consider Example 3.9 again. The function min is invertible—C3 holds for conditional possibility where conditioning is determined by min—yet, in that example, we have $Ind_s(A, B|C)$, but not $Ind_s(B, A|C)$. To get symmetry, we actually need the following variant of C3$^=$:

**C4$^=$.** If $\text{Pl}(A \cap B|C) = \text{Pl}(A' \cap B'|C')$ and $\text{Pl}(A|C) = \text{Pl}(A'|B', C') > \bot$, then $\text{Pl}(A|B, C) = \text{Pl}(A'|C')$.

**Proposition 4.2:** *If* Pl *satisfies C2$^=$ and C4$^=$, then* Pl $\models Ind_s(A, B|C)$ *if and only if* Pl $\models Ind_s(B, A|C)$.

We note that C4$^=$ is satisfied by the conditional plausibility measure $S_0$ constructed in Proposition 3.8, as well as by probability theory and $\kappa$-ranking. As Example 3.9 shows, it is not satisfied by conditional possibility with conditioning determined by min, although it is satisfied by conditional possibility with conditioning determined by multiplication.

What we are often interested in is not just the independence of one event (set) from another, but independence among a family of events. Given a set $\mathbf{A} = \{A_1, \ldots, A_k\}$ of events, define an *atom* over $\mathbf{A}$ to be an event of the form $A_1' \cap \ldots \cap A_k'$, where $A_k'$ is either $A_k$ or its complement $\overline{A_k}$. Given three sets of events $\mathbf{A}, \mathbf{B}$, and $\mathbf{C}$, we say that Pl *makes $\mathbf{A}$ strongly independent of $\mathbf{B}$ given $\mathbf{C}$*, denoted Pl $\models IND_s(\mathbf{A}, \mathbf{B}|\mathbf{C})$, if, for all atoms $X$ over $\mathbf{A}$, all atoms $Y$ over $\mathbf{B}$, and all atoms $Z$ over $\mathbf{C}$, we have Pl $\models Ind_s(X, Y|Z)$. We remark that although this definition focuses on atoms, by part (c) of Proposition 4.2, we can extend these independence assertions about atoms to independence assertions about arbitrary sets in decomposable structures. Notice that if $\mathbf{A}$ is the singleton $\{A\}$, $\mathbf{B}$ is the singleton $\{B\}$, and $\mathbf{C}$ is the singleton $\{C\}$ and we are dealing with a plausibility measure that is in fact a probability function, then this definition amounts to saying that $A$ is (probabilistically) independent of $B$ given both $C$ and $\overline{C}$.[8]

Once we consider sets of events in this way, we can consider

---

[8]This definition of independence is related to independence among *random-variables* [Pearl 1988]. Essentially, we are treating each event $A$ as a two-valued random variable that has value 1 in all worlds $w \in A$ and value 0 in worlds $w \notin A$. The following discussion can be easily extended to deal with many-valued random variables. Because of space constraints we defer that to the full version of the paper.

the *semi-graphoid* properties [Pearl 1988]:

**G1.** $IND_s(\mathbf{A}, \mathbf{B}|\mathbf{C})$ implies $IND_s(\mathbf{B}, \mathbf{A}|\mathbf{C})$.

**G2.** $IND_s(\mathbf{A}, \mathbf{B} \cup \mathbf{D}|\mathbf{C})$ implies $IND_s(\mathbf{A}, \mathbf{B}|\mathbf{C})$.

**G3.** $IND_s(\mathbf{A}, \mathbf{B} \cup \mathbf{D}|\mathbf{C})$ implies $IND_s(\mathbf{A}, \mathbf{B}|\mathbf{C} \cup \mathbf{D})$.

**G4.** $IND_s(\mathbf{A}, \mathbf{B}|\mathbf{C})$ and $IND_s(\mathbf{A}, \mathbf{D}|\mathbf{C} \cup \mathbf{B})$ implies $IND_s(\mathbf{A}, \mathbf{B} \cup \mathbf{D}|\mathbf{C})$.

These properties are well-known to hold for probabilistic independence [Pearl 1988]. They also hold in the minimal decomposable conditional plausibility measures of Proposition 3.8. In fact, we can prove a somewhat stronger result.

**Theorem 4.3:** *If* Pl *is a conditional plausibility measure satisfying* $C2^=$, $C4^=$, *and* $DECOMP_c^=$, *then* $IND_s$ *satisfies the semi-graphoid properties in* Pl.

This result is somewhat similar to one of Wilson [1994], and shows that we need relatively little to get a non-probabilistic notion of independence that obeys all the semi-graphoid axioms.

It is known that any collection of independence statements that satisfy the semi-graphoid properties can be represented using a Bayesian network [Pearl 1988]. As we just saw, it takes fairly little to get a notion of plausibilistic independence that obeys all the semi-graphoid properties. Darwiche [1992] shows how to construct Bayesian networks in the framework of [Darwiche and Ginsberg 1992]. As we observed, this framework embodies additional assumptions beyond C1, $C2^=$, $C4^=$, and $DECOMP_c^=$. While this extra structure, particularly properties like invertibility and commutativity, may not be necessary to define Bayesian networks, Darwiche's results suggest that they may be useful in facilitating analogues of probabilistic algorithms for updating beliefs in Bayesian networks. Whether this structure really is necessary is an issue that deserves further exploration.

We now turn to a more qualitative notion of independence. If **A** is strongly independent of **B** given **C**, then discovering that some atom over **B** holds does not change the plausibility of atoms over **A**. If all we care about is the relative plausibility of events, then we can consider a weaker notion, where discovering that some atom over **B** holds does not change the relative plausibility of atoms over **A**. Thus, we say Pl *makes* **A** *weakly independent of B given C*, denoted $\text{Pl} \models Ind_w(\mathbf{A}, \mathbf{B}|\mathbf{C})$, if either $\text{Pl}(B \cap C) = \bot$ or for all atoms $X$ and $X'$ over **A**, we have $\text{Pl}(X|C) \leq \text{Pl}(X'|C)$ if and only if $\text{Pl}(X|B \cap C) \leq \text{Pl}(X'|B \cap C)$. We say Pl *makes* **A** *weakly independent of* **B** *given* **C**, denoted $\text{Pl} \models IND_w(\mathbf{A}, \mathbf{B}|\mathbf{C})$, if, for all atoms $Y$ over **B** and all atoms $Z$ over **C**, we have $\text{Pl} \models Ind_w(\mathbf{A}, Y|Z)$. It is easy to see that $IND_s(\mathbf{X}, \mathbf{Y}|\mathbf{Z})$ implies $IND_w(\mathbf{X}, \mathbf{Y}|\mathbf{Z})$, as the names suggest. There do not seem to be any straightforward conditions that force weak independence to obey the semi-graphoid properties. Nevertheless, as we shall see below, it may be that weak independence has an important role to play in default reasoning.

**4.2 QUALITATIVE REASONING**

We briefly review some of the results on default reasoning from [Friedman and Halpern 1995a; Friedman et al. 1995], in light of the results of the previous sections.

A default is a formula of the form $\varphi \rightarrow \psi$, where $\varphi$ and $\psi$ are propositional formulas. Such a default is read "$\varphi$'s are typically (or normally, or by default) $\psi$'s". For the discussion here, we identify a formula $\varphi$ with the set consisting of the worlds where $\varphi$ is true; this allows us to work with sets rather than formulas. We then say that a default $A \rightarrow B$ is *satisfied* by a a plausibility measure Pl if either $\text{Pl}(A) = \bot$ or $\text{Pl}(A \cap B) > \text{Pl}(A \cap \overline{B})$, i.e., either $A$ is implausible and we accept the default vacuously, or $B$ is more plausible than its complement given $A$. This semantics for defaults is identical to that given for possibility measures [Benferhat et al. 1992] and for ordinal ranking structures [Goldszmidt and Pearl 1992]. As shown in [Friedman and Halpern 1995a], we can map other semantic structures used for giving semantics to defaults—including preferential structures [Kraus et al. 1990] and the *parameterized probability distributions* (PPDs) used in $\epsilon$-semantics [Goldszmidt et al. 1993]—to plausibility structures in such a way as to preserve the semantics of defaults.

Default reasoning in all these approaches is characterized by a collection of properties known as the KLM axioms [Kraus et al. 1990]. Our semantics for defaults does *not* guarantee that the KLM properties are satisfied. (In particular, they are not satisfied by probability measures.) What extra conditions do we have to place on plausibility measures to ensure that these properties are satisfied? In the presence of A1, the following axioms turn out to be all that we need

**A2.** If $A$, $B$, and $C$ are pairwise disjoint sets, $\text{Pl}(A \cup B) > \text{Pl}(C)$, and $\text{Pl}(A \cup C) > \text{Pl}(B)$, then $\text{Pl}(A) > \text{Pl}(B \cup C)$.

**A3.** If $\text{Pl}(A) = \text{Pl}(B) = \bot$, then $\text{Pl}(A \cup B) = \bot$.

In [Friedman and Halpern 1995a] we prove that a plausibility measure satisfies all the KLM properties if and only if it satisfies A2 and A3. Thus, A2 and A3 capture the essence of the KLM properties.

Since all the plausibility structures that arise from possibility measures, $\kappa$-rankings, preference structures, and PPDs, satisfy A2 and A3, this explains why the KLM properties hold in these approaches.[9] Moreover, as we said in the introduction, in [Friedman and Halpern 1995a] we also describe a weak necessary and sufficient condition for the KLM properties to be complete. This condition is easily seen to hold in all of these approaches. These results help us understand why the KLM properties characterize default reasoning in several different approaches. Plausibility structures also give us the tools to show how these approaches all differ once we allow first-order defaults [Friedman et al. 1995].

While there is general agreement that the KLM proper-ties are the "core" of default reasoning, there is also general agreement that they are too weak. For example suppose that a knowledge base contains only the default $Bird \rightarrow Fly$. Using the KLM properties, we cannot derive $Bird \land Red \rightarrow Fly$, since it is consistent with this knowledge base that red birds are exceptional. Yet, given that our knowledge base does not contain any information indicating that red birds are exceptional, we would like to infer that they are not. This problem has been dubbed the *irrelevance problem*, since we want our inference procedure to consider $Red$ irrelevant to $Bird \rightarrow Fly$. A great deal of recent work on default reasoning (e.g., [Goldszmidt et al. 1993; Goldszmidt and Pearl 1992; Pearl 1989]) has attempted to deal with the irrelevance problem. Roughly speaking, given a particular knowledge base, these approaches focus on a set of *preferred structures* that are determined by this particular knowledge base. Intuitively, these preferred structures satisfy the knowledge base and some additional irrelevance properties that we want to view as true by default.

---

[9] We immediately get A3, using Lemma 3.3, as a consequence of the decomposability of these approaches.

The results of the previous sections that plausibility may provide us with a general approach for dealing with irrelevance. How can we capture irrelevance? Perhaps the simplest definition is the following: $B$ is *irrelevant* to the default $C \rightarrow A$ according to Pl if $C \cap B \rightarrow A$ is satisfied by Pl if and only if $C \rightarrow A$ is satisfied by Pl. It is easy to verify that $Pl \models Ind_w(\{A\}, B|C)$ if and only if $B$ is irrelevant to $C \rightarrow A$ according to Pl. Thus, the weak notion of independence corresponds directly to irrelevance in default reasoning. Since all that matters in the semantics of a default such as $C \rightarrow A$ is the relative plausibility of the sets $C \cap A$ and $C \cap \overline{A}$, it is clear that we do not need the full power of strong independence here.

We believe that by studying $Ind_w$ in the context of default reasoning we can gain some insight into the irrelevance problem. For example, we can make precise approaches that make a maximal number of independence assumptions that are consistent with the knowledge base. We note that a similar suggestion is made by Dubois et. al. [1994] using a slightly different notion of irrelevance. They say that $B$ is irrelevant to $C \rightarrow A$ if both $C \rightarrow A$ and $C \cap B \rightarrow A$ hold. We believe that this condition is too strong, since it presupposes the acceptance of the default. We are currently exploring this issue further.

Qualitative measures also appear in the investigation of *belief change*. The problem here is how an agent should change his beliefs after making a (possibly surprising) observation. It turns out that we can use conditioning in plausibility measures to handle belief change. Roughly, we say that an agent believes $A$ given evidence $E$ if $Pl(A|E) > Pl(\overline{A}|E)$. If we want the agent's beliefs to be closed under conjunction—if the agent believes $A$ and $B$, then he also believes $A \cap B$—then we must limit our attention to qualitative plausibility measures.

Independence also proves to be important in belief change. In [Friedman and Halpern 1995b], we describe belief change in a system by putting a plausibility measure on *histories* or *runs* that describe how the system changes over time. When doing probabilistic reasoning about such systems, in many cases it is reasonable to assume that a particular state transition is independent of when it occurs. This *Markov assumption* greatly simplifies reasoning about complicated systems, and has been shown to be widely applicable in practice. There is no difficulty stating the Markov assumption using plausibilistic notions of independence. Moreover, as shown in [Friedman and Halpern 1995b], by combining the Markov assumption with qualitative plausibility measures, we get a natural and powerful model of belief change. In this model, after making an observation, the agent may revise earlier beliefs, as is done by *belief revision* [Alchourrón et al. 1985], may consider it possible that his earlier beliefs were correct but the world has changed, as is done by *belief update* [Katsuno and Mendelzon 1991], or take some combination of the two possibilities.

## 5 DISCUSSION

In this paper, we have attempted to provide an introduction to plausibility measures, with a focus on their algebraic structure. Among other things, we showed how starting with a plausibility on worlds, we can extend to a conditional plausibility measure defined on arbitrary sets, where the plausibilities have many of the properties satisfied by probability. In particular, decomposition is determined by a total function $\oplus$ with many of the key properties of addition, and conditioning is determined by a total function $\otimes$ with many of the key properties of multiplication. Moreover, the independencies in this minimal conditional plausibility space satisfy the semi-graphoid conditions. Thus, we provide a construction that can generate a Bayesian network from a preferential structure. We believe that this might have interesting implications in investigating the irrelevance problem in default reasoning.

More generally, as we hope our results show, plausibility measures give us a general framework in which to study fundamental issues of reasoning about uncertainty. They allow us to compare various approaches, and extract the key features needed to enable certain types of reasoning to be carried out. Plausibilities measures have already proved their usefulness in the analysis of qualitative default reasoning. We expect that many other applications will be found in the future.

### Acknowledgements

The authors are grateful to Adnan Darwiche, Moises Goldszmidt, and Daphne Koller for useful discussions relating to this work. The first author was supported in part by Rockwell Science Center in Palo Alto.

### References

Alchourrón, C. E., P. Gärdenfors, and D. Makinson (1985). On the logic of theory change: partial meet functions for contraction and revision. *Journal of Symbolic Logic 50*, 510–530.

Benferhat, S., D. Dubois, and H. Prade (1992). Representing default rules in possibilistic logic.

184    Friedman and HalpernIn B. Nebel, C. Rich, and W. Swartout (Eds.), *Proc. Third International Conference on Principles of Knowledge Representation and Reasoning (KR '92)*, pp. 673–684. San Francisco: Morgan Kaufmann.

Darwiche, A. (1992). *A Symbolic Generalization of Probability Theory*. Ph. D. thesis, Stanford University.

Darwiche, A. and M. L. Ginsberg (1992). A symbolic generalization of probability theory. In *Proc. National Conference on Artificial Intelligence (AAAI '92)*, pp. 622–627. Menlo Park, Calif.: AAAI Press.

Dubios, D., L. Farinãs del Cerro, A. Herzig, and H. Prade (1994). An ordinal view of independence with applications to plausible reasoning. In R. López de Mántara and D. Poole (Eds.), *Proc. Tenth Conference on Uncertainty in Artificial Intelligence (UAI '94)*, pp. 195–203. San Francisco: Morgan Kaufmann.

Dubois, D. (1986). Belief structures, possibility theory and decomposable confidence measures on finite sets. *Computers and Artificial Intelligence 5*, 403–416.

Dubois, D. and H. Prade (1990). An introduction to possibilistic and fuzzy logics. In G. Shafer and J. Pearl (Eds.), *Readings in Uncertain Reasoning*. San Francisco: Morgan Kaufmann.

Fine, T. L. (1973). *Theories of Probability*. New York: Academic Press.

Friedman, N. and J. Y. Halpern (1995a). Plausibility measures and default reasoning. Technical Report 9959, IBM. Available by anonymous ftp from starry.stanford.edu/pub/nir or via WWW at http://robotics.stanford.edu/users/nir.

Friedman, N. and J. Y. Halpern (1995b). A qualitative Markov assumption and its implications for belief change. Unpublished manuscript. Available by anonymous ftp from starry.stanford.edu/pub/nir or via WWW at http://robotics.stanford.edu/users/nir.

Friedman, N., J. Y. Halpern, and D. Koller (1995). Conditional first-order logic revisited. Unpublished manuscript. Available by anonymous ftp from starry.stanford.edu/pub/nir or via WWW at http://robotics.stanford.edu/users/nir.

Goldszmidt, M., P. Morris, and J. Pearl (1993). A maximum entropy approach to nonmonotonic reasoning. *IEEE Transactions of Pattern Analysis and Machine Intelligence 15*(3), 220–232.

Goldszmidt, M. and J. Pearl (1992). Rank-based systems: A simple approach to belief revision, belief update and reasoning about evidence and actions. In B. Nebel, C. Rich, and W. Swartout (Eds.), *Proc. Third International Conference on Principles of Knowledge Representation and Reasoning (KR '92)*, pp. 661–672. San Francisco: Morgan Kaufmann.

Greco, G. H. (1987). Fuzzy integrals and fuzzy measures with their values in complete lattices. *Journal of Mathematical Analysis and Applications 126*, 594–603.

Katsuno, H. and A. Mendelzon (1991). On the difference between updating a knowledge base and revising it. In J. A. Allen, R. Fikes, and E. Sandewall (Eds.), *Proc. Second International Conference on Principles of Knowledge Representation and Reasoning (KR '91)*, pp. 387–394. San Francisco: Morgan Kaufmann.

Kraus, S., D. J. Lehmann, and M. Magidor (1990). Nonmonotonic reasoning, preferential models and cumulative logics. *Artificial Intelligence 44*, 167–207.

Kreps, D. (1988). *Notes on the Theory of Choice*. Boulder, Colorado: Westview Press.

Pearl, J. (1988). *Probabilistic Reasoning in Intelligent Systems*. San Francisco, Calif.: Morgan Kaufmann.

Pearl, J. (1989). Probabilistic semantics for nonmonotonic reasoning: A survey. In R. J. Brachman, H. J. Levesque, and R. Reiter (Eds.), *Proc. First International Conference on Principles of Knowledge Representation and Reasoning (KR '89)*, pp. 505–516. Reprinted in *Readings in Uncertain Reasoning*, G. Shafer and J. Pearl (eds.), Morgan Kaufmann, San Francisco, Calif., 1990, pp. 699–710.

Savage, L. J. (1954). *Foundations of Statistics*. New York: John Wiley & Sons.

Shafer, G. (1976). *A Mathematical Theory of Evidence*. Princeton, N.J.: Princeton University Press.

Spohn, W. (1987). Ordinal conditional functions: a dynamic theory of epistemic states. In W. Harper and B. Skyrms (Eds.), *Causation in Decision, Belief Change and Statistics*, Volume 2, pp. 105–134. Dordrecht, Holland: Reidel.

Wang, Z. and G. J. Klir (1992). *Fuzzy Measure Theory*. New York: Plenum.

Weber, S. (1991). Uncertainty measures, decomposability and admissibility. *Fuzzy Sets and Systems 40*, 395–405.

Weydert, E. (1994). General belief measures. In R. López de Mántara and D. Poole (Eds.), *Proc. Tenth Conference on Uncertainty in Artificial Intelligence (UAI '94)*, pp. 575–582. San Francisco: Morgan Kaufmann.

Wilson, N. (1994). Generating graphoids from generalized conditional probability. In R. López de Mántara and D. Poole (Eds.), *Proc. Tenth Conference on Uncertainty in Artificial Intelligence (UAI '94)*, pp. 583–591. San Francisco: Morgan Kaufmann.